# Dynamical Mode Recognition of Coupled Flame Oscillators by Supervised and Unsupervised Learning Approaches

Weiming Xu [a,#], Tao Yang [b,#], and Peng Zhang [a,*]

a Department of Mechanical Engineering, City University of Hong Kong, Kowloon Tong, Kowloon, 999077, Hong Kong
b Department of Mechanical Engineering, The Hong Kong Polytechnic University, Hung Hong, Kowloon, 999077, Hong Kong

**Abstract**

Combustion instability in gas turbines and rocket engines, as one of the most challenging problems in combustion research, arises from the complex interactions among flames influenced by chemical reactions, heat and mass transfer, and acoustics. Identifying and understanding combustion instability is essential for ensuring the safe and reliable operation of many combustion systems, where exploring and classifying the dynamical behaviors of complex flame systems is a core task. To facilitate fundamental studies, the present work concerned dynamical mode recognition of coupled flame oscillators made of flickering buoyant diffusion flames, which have gained increasing attention in recent years but are not sufficiently understood. The time series data of flame oscillators were generated through fully validated reacting flow simulations. Due to the limitations of expertise-based models, a data-driven approach was adopted. In this study, a nonlinear dimensional reduction model of variational autoencoder (VAE) was used to project the high dimensional data onto a 2-dimensional latent space. Based on phase trajectories in the latent space, both supervised and unsupervised classifiers were proposed for datasets with and without well-known labeling, respectively. For labeled datasets, we established the Wasserstein-distance-based classifier (WDC) for mode recognition; for unlabeled datasets, we developed a novel unsupervised classifier (GMM-DTW) combining dynamic time warping (DTW) and Gaussian mixture model (GMM). Through comparing with conventional approaches for dimensionality reduction and classification, the proposed supervised and unsupervised VAE-based approaches exhibit a prominent performance across seven assessment metrics for distinguishing dynamical modes, implying their potential extension to dynamical mode recognition in complex combustion problems.

---

[*] Corresponding author
  E-mail address: penzhang@cityu.edu.hk, Tel: (852)34429561.
[#] The authors equally contributed to the work.

# 1. Introduction

Diffusion flames are ubiquitous in nature (e.g., wildland and urban fires), domestic applications (e.g., fireplaces and furnaces), and industrial applications (e.g., gas turbines and rocket engines). In a diffusion flame, the fuel and oxidizer are initially and spatially separated, so it is often referred to as a non-premixed flame. These combustion phenomena involve complex processes of fluid mechanics, chemical reaction kinetics, thermodynamics, etc., with outputs influenced by factors such as the composition and properties of the fuel, the availability of the oxidizing agent (usually oxygen), and the reaction conditions (e.g., temperature and pressure). Understanding and controlling these factors is crucial for optimizing combustion processes and minimizing environmental impacts. Previously, the control and monitoring of combustion instabilities were achieved through combustion system designing and modeling [1]. However, as the complexity of combustion systems increases, the difficulty of building the combustion models increases dramatically.

A flickering flame, as a classical unsteady flame configuration, was experimentally observed as the vibratory motion of Bunsen diffusion flames (referred to as "the flicker of luminous flames") by Chamberlin and Rose [2]. They found that "the upper portion of the luminous zone rises to a maximum height ten times per second", and that "this rate of vibration is not greatly affected by change in conditions". Such a flame vibration was also observed in pool fires (puffing fires) [3] and premixed flames [4, 5]. In recent years, coupled flickering diffusion flames have increasingly gained attention, as larger systems of flickering buoyant diffusion flames give rise to richer dynamical phenomena. For example, Kitahata et al. [6] reported that two identical oscillating candle flames exhibit in-phase and anti-phase modes by increasing the distance between the flames. Okamoto et al. [7] observed four distinct dynamical modes of three flickering candle flames in an equilateral triangle arrangement, including the in-phase, partial in-phase, rotation, and death modes. Manoj et al. [8] experimentally found variants of clustering and chimera states of 5-7 candle-flame oscillators in annular networks.

Flame mode recognition plays a crucial role in enhancing the understanding, modeling, and application of complex combustion systems. Identifying flame modes is essential to understanding flame characteristics under varying conditions [9], informing theoretical modeling and engineering practices. The identified modes can facilitate the construction of reduced-order models for dynamical flame behaviors, aiding in the prediction of flame stability and combustion efficiency. Generally, the reduced models can provide crucial insights for designing and optimizing systems of flow and combustion [10, 11]. Beyond the practical applications, flame mode recognition also delves into fundamental scientific principles. Investigating the formation mechanisms and dynamical characteristics of different flame modes is beneficial to advance the theoretical insights of complex physical phenomena in combustion processes. This exploration can drive cutting-edge methods for monitoring and controlling practical flames [12].

However, conventional knowledge-based approaches for mode recognition of coupled flickering flames have faced challenges, particularly when the complexity of a flame system increases. Challenges arise with phenomena such as large flame groups, which pose obstacles in feature extraction through expert knowledge, as well as strong turbulence and intense interactions between chemistry and flow. Therefore, the present study aims to develop deep learning-based methods for dynamical mode recognition of coupled flame oscillators made of identical flickering flames.

The first challenge of the study is to deal with the intrinsically infinite dimension of flames, which are described by a set of partial differential equations accounting for the conservation laws. To enable latent features within the lower dimensional space, the primary approach involves identifying an effective low-dimensional latent representation of high-dimensional dynamical data [13-15]. Many linear reduction techniques, such as principal component analysis (PCA) [16] and proper orthogonal decomposition (POD) [17], have been utilized in extensive applications. However, the linear nature of these approaches may lead to suboptimal dimensionality reduction as the complexity of the nonlinear system increases [18]. Recently, deep learning (DL) has emerged as a prominent tool and has been applied to various combustion issues [19], for example, data analysis, mode recognition, and physical modeling.

Nonlinear dimensionality reduction methods have gained significant interest and been gradually used in combustion applications. Variational autoencoder (VAE) is one of the most advanced approaches in the field of flame dynamics. Recently, Arai et al. [20] used a VAE to carry out the nonlinear reduction of various physics quantities in combustion systems. Iemura et al. [21] combined VAE and POD to analyze the flame oscillation phenomenon in the low-dimensional space. Considering the hidden states of complex combustion through time, Xu et al. [22] proposed a two-layer bidirectional long short-term memory variational autoencoder (Bi-LSTM-VAE) model for dynamical mode recognition of multiple-flame systems. By classifying the low-dimensional trajectories of different dynamical processes, combustion states can be distinguished and recognized well.

Based on the adequacy of labeling in the dataset, mode recognition methods are generally categorized into supervised, semi-supervised, and unsupervised types. The first two methods play an important role in analyzing sufficient labeled data. Liu et al. [23] employed the multi-classified support vector machine (SVM) algorithm for the supervised automatic classification of various combustion modes. Wang et al. [24] introduced a supervised deep learning approach based on convolutional neural networks (CNN) and deep neural networks (DNN) for monitoring combustion status and predicting heat release rates. In those studies, ample data labels are crucial for achieving optimal performance, especially in the cases of supervised methods. In practice, datasets with labels are rare and the obtaining process is costly and poses significant challenges. Therefore, unsupervised clustering methods are extensively employed for complex combustion issues [19]. Among them, time series clustering is a type of data analysis technique for revealing the inherent structures and patterns in time-dependent data.

Many previous studies found that time series clustering is highly effective for datasets with strong spatial-temporal characteristics. In fluid mechanics, Liu et al. [25] calculated the clustering index of wind turbines within a wind farm based on dynamic time warping (DTW) and implemented adaptive and unsupervised turbine clustering using the Gaussian mixture model (GMM). Barwey et al. [26] proposed a temporal axis clustering approach to enable mode recognition of similar regions of unsteady flow. In combustion, Mishra et al. [27] developed a novel hybrid unsupervised cluster-wise regression approach to represent the flamelet tables and speed up the computations in turbulent combustion simulations. Castellanos et al. [28] applied a $k$-means clustering algorithm to group combustion chemical features for constructing a lower-order predictive model.

Despite the noteworthy studies in dimensional reduction and mode recognition, in discovering dynamical behaviors of coupled flickering diffusion flames, there are many interesting problems to be solved and the study is still in its infant stage. Consequently, the present study was prompted by recognizing two key deficiencies in existing experimental studies and data analysis approaches. Firstly, traditional methodologies struggle to recognize modes within high-dimensional complex systems due to the challenge of distilling features from expansive dynamical frameworks and discerning dynamic modes amidst the intricate interplay of physics and chemistry coupling. Chi et al. [29] proposed a state-of-the-art physical-model-based method for flame dynamical mode recognition. This method first extracts three-dimensional flame features from infinite-dimensional flame snapshots based on expert knowledge and then identifies dynamical flame modes by calculating the Wasserstein distance of phase trajectories. While their physical model is commendable for flame mode recognition, expert-driven feature extraction faces significant challenges in complex flame systems, limiting its robustness and scalability. Secondly, a lack of robust and comprehensive frameworks for mode recognition persists, both in labeled and unlabeled datasets of coupled flickering flames.

In this study, the datasets consist of time series signals from coupled flickering flame systems, generated through rigorous numerical simulations. Each mode within the datasets represents a carefully validated state of the combustion system, ensuring the reliability and accurate representation of the underlying physical phenomena. Both supervised and unsupervised mode recognition methods will be proposed for the presence or absence of known labeling in the dataset, respectively. First, this study will employ the nonlinear dimensionality reduction of VAE to obtain a two-dimensional phase trajectory of latent variables, which no longer depends on expert knowledge but rather harnesses deep learning methods. Second, supervised and unsupervised classifiers will be utilized to analyze the lower dimensional phase trajectories. For the labeled datasets, the Wasserstein distance (WD) will be used to quantify distributional similarities of labeled and unlabeled phase trajectories for dynamic flame mode recognition. To deal with unlabeled time series data, this study will combine a dynamic time warping (DTW) method and an unsupervised Gaussian mixture model (GMM) for effectively identifying the dynamic flame modes. The present study mainly focuses on the establishment of a supervised and unsupervised learning framework for the dynamical mode recognition of combustion systems, such as the coupled flame systems of dual and triple flickering buoyant diffusion flames, providing a potential approach to more complex systems with additional flame oscillators and real combustion systems.

## 2. Deep Learning Methodology

Traditional methods encounter formidable challenges in mode recognition within high-dimensional complex systems, grappling with the effective extraction of features from large-scale dynamical systems and the discernment of dynamic modes amidst the intricate interactions between physics and chemistry coupling. Deep learning-based approaches have emerged as effective and alternative methodologies to address these challenges. In this study, two supervised and unsupervised learning approaches for mode recognition of dynamically coupled flickering flames will be developed based on the dimensionality reduction of VAE.

The detailed structures of the designed VAE-WD and VAE-GMM-DTW approaches are shown in Fig. 1. The present datasets consist of $m$ features and $n$ samples from numerical simulations of flickering buoyant diffusion flames. Flame data are reconstructed through the four-layer encoder, two-dimensional latent representation, and four-layer decoder of VAE. For the phase trajectory of latent variables, the Wasserstein-distance-based supervised classifier and the GMM-DTW unsupervised classifier are employed to identify various dynamical modes of coupled flickering flames. The present approaches will be elaborated in the following subsections.

### 2.1 Variational Autoencoder (VAE)

Linear dimensional reduction models like PCA, POD, and DMD have been employed to extract dominant spatial-temporal features in combustion research. However, analyzing coupled flame oscillators poses a formidable challenge due to the intricate interplay of spatial-temporal coherent structures in high-dimensional nonlinear dynamics. Traditional linear methods struggle to address this complexity because their inherent linearity leads to performance degradation in nonlinear problems, requiring an excessive number of modes to achieve the same reconstruction accuracy as nonlinear dimensional reduction methods. Deep neural networks have emerged as an alternative to linear methods for their effectiveness in handling highly nonlinear problems. Hence, in this subsection, we describe a method for efficiently transforming high-dimensional trajectories into a simplified, low-order phase space. This is accomplished by employing a variational autoencoder, which holds great potential for application in various complex nonlinear systems.

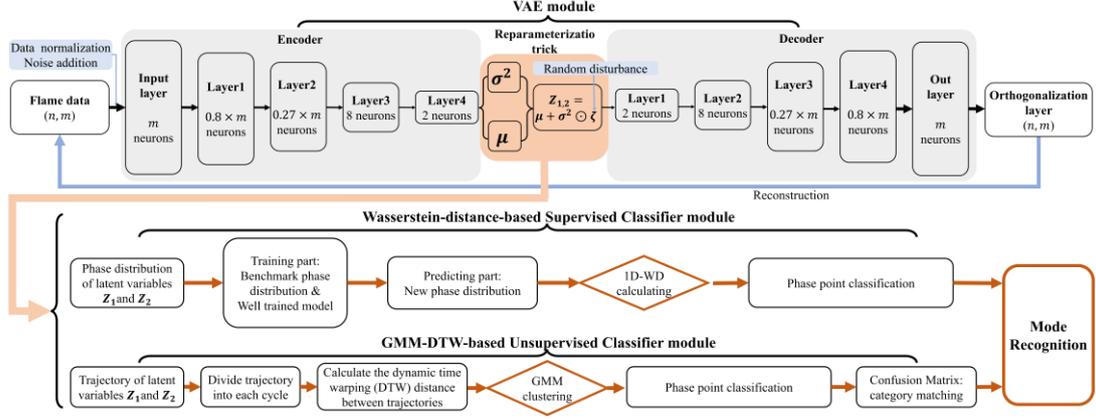

Fig. 1. Structure of the designed variational autoencoder with Wasserstein distance (VAE-WD) and Gaussian mixture model and dynamic time warping (VAE-GMM-DTW) approaches consisting of two modules: a neural network model of VAE and flowcharts of the supervised Wasserstein-distance-based classifier and the unsupervised GMM-DTW-based classifier.

Variational autoencoder (VAE), a powerful deep generative model, has been widely utilized to analyze the low-dimensional latent features of high-dimensional complex data. As shown in Fig. 2, the nonlinear dimensionality reduction and reconstruction are carried out through the multiple neural network layers of the encoder and decoder modules. The time series of physical quantities in combustion systems are collected from multiple sensors distributed along each flame. The data collection will be given in detail in Section 3.

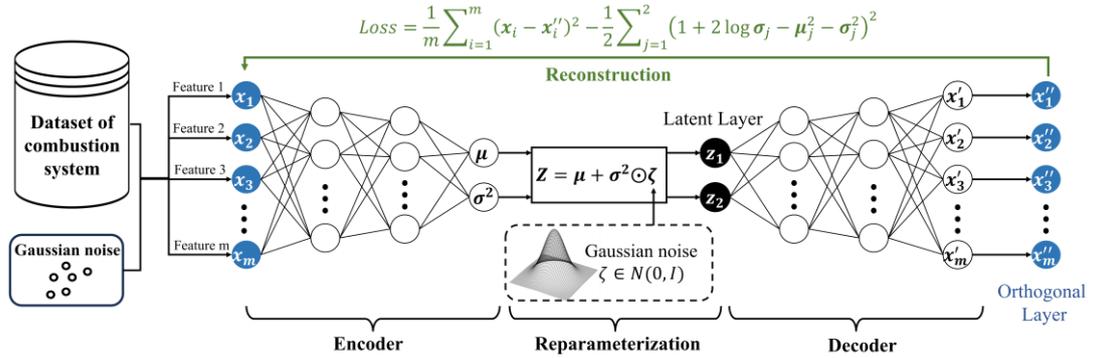

Fig. 2. Data flow through the present VAE model: dataset, encoder, reparameterization, decoder, and orthogonal layer.

Here, a concise sketch of the present VAE model is given, while the detailed formulations are presented in the Supporting Material. In the encoder component of the VAE, the feature vectors ($m$ dimensions) are nonlinearly mapped by multiple neural network layers with decreasing numbers of neurons. The output of the four-layer encoder connects to a latent space (also called representation space) of two variables. In other words, those physical features can be encoded to a phase point of the latent space and the time series data can form a continuous phase trajectory in the two-dimensional phase plane. Then, the decoder component will reconstruct the physical features from the two latent variables through the multiple neural network layers with increasing numbers of neurons. Finally, an additional layer following the decoder is devised to orthogonalize the decoder output vector $X'$ into the orthogonal output vector $X''$. The objective of the orthogonal layer is to minimize the mean square error between the reconstructed output and the original input. This structure contributes to a more stable and efficient optimization process, accelerating training convergence and enhancing overall learning performance. Simultaneously, a Kullback-Leibler (KL) divergence term is used to ensure that the phase distribution of latent variables aligns with a predefined prior distribution, namely the Gaussian distribution. Specifically, the objective of the KL divergence is to maximize the Evidence Lower Bound (ELBO) to provide a lower bound to the marginal likelihood $P(x)$, which is formulated as

$$\log(P(x)) = E_{z \sim q_\theta}[\log p(x|z)] - KL(q_\theta(Z|X)||p(Z)) \tag{1}$$

where

$$E_{z \sim q_\theta}[\log p(X|Z)] = \frac{1}{m} \sum_{i=1}^{m} (x_i - x_i'')^2 \tag{2}$$

represents the reconstruction error between the input $X = \{x_1, x_2, \cdots, x_m\}$ and output $X'' = (x_1'', x_2'', x_3'', \cdots, x_m'')$, and

$$KL(q_\theta(Z|X)||p(Z) = \frac{1}{2} \sum_{j=1}^{2} (1 + 2\log \sigma_j - \mu_j^2 - \sigma_j^2) \tag{3}$$

represents the regularization term of the posterior distribution $q_\theta(Z|X)$ and the prior distribution $p(z) \sim N(0, I)$.

Particularly, the VAE model incorporates a reparameterization trick to handle the sampling from $q_\theta(Z|X)$ during gradient-based optimization. When a stochastic variable $Z$ is sampled from a distribution $N(\mu, \sigma^2)$, where $\mu$ and $\sigma^2$ are directly sampled randomly, the neural network cannot calculate the gradient during backward optimization, thus hindering parameter optimization [30]. The reparameterization trick can overcome this difficulty by introducing a standard normal distribution and a random sampling for the distribution of the stochastic variable $Z$. Specifically, the stochasticity of the latent variable $Z = (Z_1, Z_2)$ is expressed by

$$Z = \mu + \sigma^2 \odot \zeta \tag{4}$$

where $\zeta \sim N(0, I)$ is a Gaussian noise with mean value 0 and covariance matrix $I$ (the identity matrix); $\odot$ denotes the Hadamard product. By reparametrizing the stochastic variable $Z$ into a deterministic variable dependent on $\mu$ and $\sigma$, the network can effectively optimize parameters through the sampling process.

The input $X = \{x_1, x_2, \cdots, x_m\}$ and the output $Y = \{\mu, \sigma^2\}$ of the encoder are formulated as

$$Y = f_E(W_E X + b_E) \tag{5}$$

where $W_E$ and $b_E$ respectively correspond to the weights and bias matrixes of the encoder network, $f_E$ is the activation function of the encoder, and all neural layers utilize the ReLU (rectified linear unit) activation function. In the decoder, $Z$ is reconstructed to $X' = (x_1', x_2', x_3', \cdots, x_m')$ with the same dimension of $X$. The relationship between $Z$ and $X'$ is expressed as

$$X' = f_D(W_D Z + b_D) \tag{6}$$

where $W_D$ and $b_D$ respectively correspond to the weights and bias matrixes of the decoder network, $f_D$ is the activation function of the decoder, utilizing the Sigmoid activation for the output layer, and the ReLU activation for all other layers. In addition, to enhance the better training dynamics and avoid issues like vanishing and exploding gradients from the reconstruction process, an orthogonal layer, initialized with an orthogonal weight matrix, is employed to the decoder output $X'$ and the final produce $X''$:

$$X'' = W_{orthogonal}X' + b_{orthogonal} \tag{7}$$

where $W_{orthogonal}$ indicates the weight of the last layer. The training process of the VAE is presented in Algorithm 1.

---

Algorithm 1: Training process of the VAE
---

    Model *encoder E and decoder D of the VAE.*
    Input data $X = \{x_1, x_2, \cdots, x_m\}$ is split into $X_{train}$ and $X_{val}$.
    **repeat:**
   **for** $x$ **in** $X_{train}$: # $x$ *represents different batches of samples.*
   $\mu, \sigma \leftarrow f_E(W_E x + b_E)$
   $z \leftarrow \mu + \sigma^2 \odot \zeta, \zeta \leftarrow N(0, I)$
   $x' \leftarrow f_D(W_D z + b_D)$
   $x'' \leftarrow W_{orthogonal} x' + b_{orthogonal}$
   $Loss = Loss_{rec} + KL \leftarrow \text{MSE}(x, x''), -\frac{1}{2}\sum_{j=1}^{2}(1 + 2\log \sigma_j - \mu_j^2 - \sigma_j^2)$
   update $W_E, b_E, W_D, b_D, W_{orthogonal}, b_{orthogonal}$ using the backpropagation algorithm.
   $Loss_{val} \leftarrow$ computed from $X_{val}$
   **if** $Loss_{val} < min(Loss_{val}.history)$ **then**
     save model, $min(Loss_{val}.history) \leftarrow Loss_{val}$
   **else** *count* += 1
     **if** *count* > *stop_limit* **then** stop training   # *stop_limit is the stop setting of early stopping strategy.*
    **until** the final epoch is reached.

---

In the present VAE model, the Adam optimization algorithm and the minibatch training method with a batch size of 64 are performed. The deep learning structure Pytorch is utilized as the backend computation library. Both the encoder and decoder components respectively consist of four layers, with $m$, $0.8 \times m$, $0.27 \times m$, 8, and 2 neurons and 2, 8, $0.27 \times m$, $0.8 \times m$, and $m$ neurons, where $m$ denotes the number of collected features. All codes are run on the Google Colab platform, utilizing a computational environment with a V100 GPU, and the training time for our two-flame cases with a data size of (8000, 160) is only 314 seconds. For the triple-flame cases with a data size of (16000, 240), the training time is 516 seconds. The software libraries and configurations used include Python version 3.10.12, CUDA version 12.1, and Pytorch version 2.3.0+cu121. For other configurations, please refer to the default setup of Google Colab. In the training process of the VAE model, an early stopping strategy is adopted, either finishing 100 iterations with no loss reduction or reaching the pre-defined epoch limit of 2000. The reconstruction loss of present training and validation data in VAE is given in Fig. S1.

It is noteworthy that the dimensionality reduction model in this study serves the purpose of achieving precise dynamical mode recognition. In the present problem of coupled flame oscillators, the two-dimensional latent space effectively captures reconstruction information, as demonstrated by the minimal reconstruction errors observed. Furthermore, the distribution of phase points for each mode within this two-dimensional latent space is distinct, suggesting a notable advantage in mode recognition. These findings will be further elaborated upon in Section 4, emphasizing the robust representational power offered by a two-dimensional phase space for dynamic data. Also, we notice that more complex combustion systems, for example, turbulent flames in annular combustors, need to be represented by more than two-dimensional latent space. A high-dimension latent space could be achieved by designing an encoder to map input data to a higher-dimensional latent space, outputting both the $\mu$ and $\sigma$ of the latent variables. Also, we optimize the VAE by minimizing a composite loss function, including the reconstruction loss and the KL divergence. The decoder then reconstructs the input data from these low-dimensional latent variables, effectively capturing complex variations in complex combustion systems.

For applications in large datasets, such as multiple flame systems or real-time complex combustion systems, the computational cost of our deep learning approaches can be reduced through: 1) Compression techniques, like quantization, which can compress weight matrices by reducing precision (e.g., from 32-bit floating point values to 8-bit integers) with minimal quality loss [31]; 2) Learning techniques, such as distillation, which can train models to mimic larger models with improving accuracy and enabling parameter reduction for a smaller footprint [32]; and 3) Automation tools, including Hyper-Parameter Optimization (HPO) and architecture search, which can optimize hyper-parameters and model architecture to enhance accuracy and efficiency with reducing the model size and latency [33].

**2.2 Variational Autoencoder with Wasserstein Distance (VAE-WD) Classifier**

Wasserstein distance (WD) or Kantorovich–Rubinstein metric, inspired by the concept of optimal transportation, is used to measure the similarity between probability distributions [34]. The Wasserstein metric has been widely applied in various mathematical fields such as fluid mechanics, partial differential equations, optimization, probability theory, and statistics [35]. Moreover, it offers a successful framework for comparing diverse objects in practical applications like image retrieval, computer vision, pharmaceutical statistics, genomics, economics, and finance [36]. Recently, Chi et al.'s research [29] demonstrated the effectiveness of using the Wasserstein distance, a distance metric between probability distributions in phase space, for flame mode recognition. However, extracting the flame brightness-based features through expert knowledge may encounter challenges as the complexity of a flame system increases, especially in the case of large flame groups, which can hinder the brightness feature extraction of individual flames.

To distinctly identify the various dynamic modes within the phase space of the nonlinear dimensionality reduction VAE model, it is crucial to ensure quantitative analysis of the mode recognition results. Inspired by this innovative approach and leveraging the capabilities of nonlinear reduced-order models (ROM), our study employs the VAE-based neural network to project high-dimensional flame data onto a low-dimensional phase space, bypassing the limitations of expert knowledge and adopting a data-driven approach. When dealing with a sufficiently labeled dataset, we utilize the WD value to evaluate the similarity among phase trajectories of flames, enabling a supervised mode recognition.

As shown in Fig. 3, the VAE-WD-based supervised classification has the training component for benchmark cases and the recognizing component for predicting cases. The first component is to train a VAE model and label the two-dimensional phase distribution of each known flame mode. The labeled phase trajectories are taken as benchmarks to classify different flame cases. The second component is to project a new set of flame data into the two-dimensional phase distribution via the well-trained VAE model and reshape the phase points into a one-dimensional sequence.

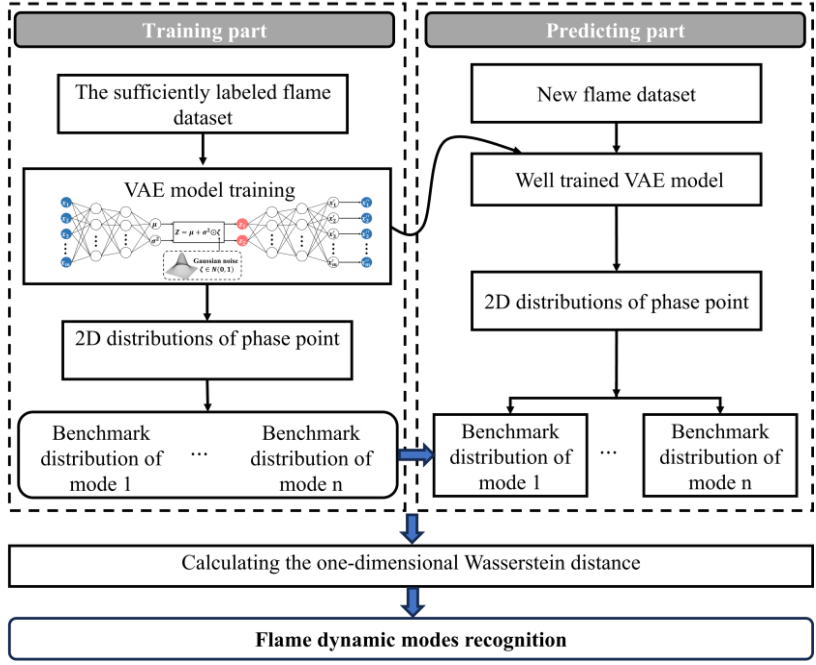

Fig. 3. Structure and flowchart of the VAE-WD classifier, including training part, predicting part, and WD-based classifier.

Specifically, the Wasserstein distance (i.e., Earth Mover's Distance [37]) is calculated to assess the similarity between the new distribution and each benchmark distribution. The one-dimensional sequence contains multiple circles. The average of WD values of all circles is obtained:

$$W(\boldsymbol{P}, \boldsymbol{Q}) = \frac{1}{N}\sum_{i=1}^{N} inf_{\gamma \epsilon \Gamma(p,q_i)} \int_{X \times Y} c(\boldsymbol{x}, \boldsymbol{y}) \, d\gamma(\boldsymbol{x}, \boldsymbol{y}) \tag{8}$$

where $\boldsymbol{P}$ and $\boldsymbol{Q}$ represent the new and benchmark probability distributions, respectively; $N$ denotes the number of circles of the benchmark; $\gamma$ is a joint distribution on $X \times Y$ with marginals $\boldsymbol{P}$ and $\boldsymbol{Q}$; $\Gamma(\boldsymbol{p}, \boldsymbol{q}_i)$ is the set of all joint distributions; $\boldsymbol{q}_i$ is one of the circles of the benchmark distribution; $X$ and $Y$ are the spaces of the $\boldsymbol{p}$ and $\boldsymbol{q}$ distributions; and $c(\boldsymbol{x}, \boldsymbol{y})$ is the cost of transporting mass from $\boldsymbol{x}$ to $\boldsymbol{y}$.

When optimizing the Wasserstein distance, we leverage the Kantorovich-Rubinstein duality theorem to transform the problem into optimizing overall 1-Lipschitz functions [38]. In other words, Eq. (8) can be expressed as

$$W(\boldsymbol{P}, \boldsymbol{Q}) = \frac{1}{N}\sum_{i=1}^{N} sup_{\|f\|_{Lip \leq 1}} \left( \int_{\chi} f(\boldsymbol{x}) dP(\boldsymbol{x}) - \int_{\chi} f(\boldsymbol{x}) dQ(\boldsymbol{x}) \right) \tag{9}$$

Here, $\|f\|_{Lip \leq 1}$ denotes that $f$ is 1-Lipschitz, satisfying $|f(\boldsymbol{x}) - f(\boldsymbol{y})| \leq d(\boldsymbol{x}, \boldsymbol{y})$ for all $\boldsymbol{x}, \boldsymbol{y} \in \chi$, where $d(\boldsymbol{x}, \boldsymbol{y})$ is the Euclidean distance function. The optimization objective is to identify a function $f$ that maximizes the above expression, thereby measuring and maximizing the difference between $\boldsymbol{P}$ and $\boldsymbol{Q}$. This method not only simplifies the challenging task of optimizing couplings but also leverages advanced mathematical optimization theories, such as convex optimization and duality theorems to effectively compute the Wasserstein distance.

**2.3 Gaussian Mixture Model and Dynamic Time Warping (GMM-DTW) Classifier**

Acquiring a thoroughly labeled dataset is often costly and challenging in industrial processes, particularly in complex combustion systems, where the scarcity of data and the complexity of structures pose significant hurdles. Consequently, the development of effective unsupervised approaches for flame dynamic mode recognition emerges as a promising avenue. These methods provide valuable insights into the behaviors of dynamic modes in multi-flame systems, especially in scenarios where data are limited and not well understood. Among these approaches, time series clustering stands out as an important method for uncovering the underlying structures and patterns in time-dependent data, particularly for effectively analyzing datasets with strong spatial-temporal characteristics. Motivated by the work of Liu et al. [21], who applied the Dynamic Time Warping (DTW) algorithm to process time series data of active power between generators and achieved accurate unsupervised clustering and classification, we utilize an unsupervised time series clustering method. This approach leverages the Gaussian Mixture Model (GMM) and DTW algorithm to capture the temporal information of continuous physical phenomena observed in each cycle of the dynamical flame system, enabling accurate mode recognition.

The machine-learning-based clustering method, GMM, has gained widespread popularity in mode recognition. GMM is a probabilistic model and consists of $K$ sub-distributions, each of which is a single Gaussian distribution [39]. The overall probability density function (PDF) of a GMM is formed by a weighted linear combination of individual Gaussian PDFs:

$$P(\mathbf{Z}) = \sum_{i=1}^{K} \boldsymbol{\omega}_i N(\mathbf{Z}|\boldsymbol{\mu}_i, \boldsymbol{\Sigma}_i) \tag{10}$$

$$N(\mathbf{Z}|\boldsymbol{\mu}_i, \boldsymbol{\Sigma}_i) = \frac{1}{\sqrt{2\pi|\boldsymbol{\Sigma}_i|}} exp\left\{-\frac{1}{2}(\mathbf{Z} - \boldsymbol{\mu}_i)^T \boldsymbol{\Sigma}_i^{-1}(\mathbf{Z} - \boldsymbol{\mu}_i)\right\} \tag{11}$$

where $N(\mathbf{Z}|\boldsymbol{\mu}_i, \boldsymbol{\Sigma}_i)$ is the probability distribution; $\mathbf{Z} = (\mathbf{Z}_1, \mathbf{Z}_2)$ is the input of GMM; $\boldsymbol{\omega}_i$ ($\sum \boldsymbol{\omega}_i = 1$), $\boldsymbol{\mu}_i$, and $\boldsymbol{\Sigma}_i$ represent the weight, mean, and covariance matrix of the $i$-th Gaussian distribution, respectively.

The GMM parameters are optimized by the expectation-maximization (EM) algorithm to maximize the likelihood of the observed data. The EM algorithm iterates between the two steps of the Expectation (E) step and the Maximization (M) step:

First, the algorithm in the E-step, computes the posterior probabilities that each data point $\mathbf{Z}_i$ is generated by each Gaussian component $K$. These probabilities are denoted as

$$\Upsilon(\mathbf{z}_{ik}) = \frac{\boldsymbol{\pi}_k N(\mathbf{Z}_i|\boldsymbol{\mu}_k, \boldsymbol{\Sigma}_k)}{\sum_{j=1}^{K} \boldsymbol{\pi}_j N(\mathbf{Z}_i|\boldsymbol{\mu}_j, \boldsymbol{\Sigma}_j)} \tag{12}$$

where $\mathbf{z}_{ik}$ is the latent variable indicating the assignment of data point $\mathbf{Z}_i$ to component $K$; $N(\mathbf{Z}_i|\boldsymbol{\mu}_k, \boldsymbol{\Sigma}_k)$ is the probability density function of the Gaussian distribution with mean $\boldsymbol{\mu}_k$ and covariance $\boldsymbol{\Sigma}_k$; and $\boldsymbol{\pi}_k$ is a priori probability for the $K$-th Gaussian component.

Second, given the probabilities in Eq. (12), the algorithm in the M-step updates the parameters $\boldsymbol{\theta} = \{\boldsymbol{\omega}_k, \boldsymbol{\mu}_k, \boldsymbol{\Sigma}_k\}$ to maximize the expected log-likelihood of the data by

$$\boldsymbol{\pi}_k = \frac{1}{N} \sum_{i=1}^{N} \Upsilon(\mathbf{z}_{ik}) \tag{13}$$

$$\boldsymbol{\mu}_k = \frac{\sum_{i=1}^{N} \Upsilon(\mathbf{z}_{ik}) \mathbf{Z}_i}{\sum_{i=1}^{N} \Upsilon(\mathbf{z}_{ik})} \tag{14}$$

$$\boldsymbol{\Sigma}_k = \frac{\sum_{i=1}^{N} \Upsilon(\mathbf{z}_{ik}) (\mathbf{Z}_i - \boldsymbol{\mu}_k)(\mathbf{Z}_i - \boldsymbol{\mu}_k)^T}{\sum_{i=1}^{N} \Upsilon(\mathbf{z}_{ik})} \tag{15}$$

Consequently, the EM algorithm can ensure that the log-likelihood increases with each iteration, leading to convergence to a local maximum of the log-likelihood function. Specifically, the E and M steps are iteratively repeated until convergence, which is typically determined by the change in the log-likelihood $\log p(\mathbf{Z}|\boldsymbol{\theta}) = \sum_{i=1}^{N} \log \left(\sum_{i=1}^{N} \boldsymbol{\pi}_k N(\mathbf{Z}_i|\boldsymbol{\mu}_k, \boldsymbol{\Sigma}_k)\right)$ of the data between iterations falling below a predefined threshold.

Dynamic time warping (DTW), a type of time-series-based distance clustering method, is used to compute the distance between two time series data, which have high similarity but do not perfectly sync up [40]. In the proposed unsupervised time series clustering method (GMM-DTW), the DTW approach is utilized to capture the temporal characteristics of the raw data before classifying samples. Specifically, the DTW distance between each circle is calculated as follows:

First, for two time series of $\boldsymbol{Q} = \{\boldsymbol{q}_1, \boldsymbol{q}_2, \cdots, \boldsymbol{q}_s\}$ and $\boldsymbol{C} = \{\boldsymbol{c}_1, \boldsymbol{c}_2, \cdots, \boldsymbol{c}_t\}$, the $d(\boldsymbol{q}_i, \boldsymbol{c}_j)$ of a $s \times t$ distance matrix $\boldsymbol{D}$ is calculated by

$$\boldsymbol{D}[i,j] = d(\boldsymbol{q}_i, \boldsymbol{c}_j) = \sqrt{(\boldsymbol{q}_i - \boldsymbol{c}_j)^2} \tag{16}$$

where $\boldsymbol{D}[i,j]$ represents the Euclidean distance between the $\boldsymbol{q}_i$ element in $\boldsymbol{Q}$ and the $\boldsymbol{c}_j$ element in $\boldsymbol{C}$.

Second, the minimum accumulated distance of the dynamic time warping is obtained by iterating

$$\boldsymbol{W}[i,j] = \boldsymbol{D}[i,j] + \min\{\boldsymbol{W}[i-1,j], \boldsymbol{W}[i,j-1], \boldsymbol{W}[i-1,j-1]\} \tag{17}$$

where $\boldsymbol{W}[i,j]$ is the accumulated distance from the start of the sequence $\boldsymbol{Q}(\boldsymbol{q}_1)$ to $\boldsymbol{Q}(\boldsymbol{q}_i)$ and from the start of the sequence $\boldsymbol{C}(\boldsymbol{c}_1)$ to $\boldsymbol{C}(\boldsymbol{c}_j)$. $\boldsymbol{W}[i-1,j]$, $\boldsymbol{W}[i,j-1]$, and $\boldsymbol{W}[i-1,j-1]$ represent the path costs from the above, left grid points, and diagonal grid points, respectively. It's worth mentioning that the warping path $\boldsymbol{W}$ is an $s \times t$ dynamic programming matrix with $\boldsymbol{W}[i,j]$ for the alignment starts from $\boldsymbol{W}[1,1] = \boldsymbol{D}[1,1]$ and ends to $\boldsymbol{W}[i,j]$. The path must satisfy constraints of boundedness, continuity, and monotonicity as well as setting all other elements in the first row and first column to infinity.

Third, after computing the dynamic programming matrix $\boldsymbol{W}$, we backtrack from the bottom-right grid point to find the optimal path, moving towards the top-left corner while selecting the path with the minimum cost at each step. This path represents the best match between the two sequences. By computing the total cost of this optimal path, we gauge the similarity between the time-series sequences, with a smaller cost indicating a higher degree of similarity.

The DTW approach has established itself as a cornerstone in various time series applications, including speech recognition, handwriting recognition, and activity recognition, owing to its adeptness in handling non-linear alignments [41]. Building upon this foundation, we present a pioneering application of DTW in recognizing the dynamical modes of coupled flame oscillators. This groundbreaking approach signifies the first instance of utilizing DTW in flame mode recognition, a domain characterized by highly nonlinear dynamics. By leveraging the flexibility of DTW, we aim to precisely distinguish the diverse dynamical modes exhibited by coupled flame oscillators, thereby advancing our understanding of these complex systems.

As shown in Fig. 4, this study incorporates the two approaches to effectively classify diverse dynamical modes of various coupled flickering flames. Given the periodic nature of present dynamical flame systems, we capitalize on this characteristic by employing the DTW distance between any two cycles to extract features from the entire trajectory of the two latent variables. This approach can capture the continuous physical phenomena observed in each cycle of the flame system by measuring alignment distances between phase point trajectories. Particularly, the newly extracted features are clustered by GMM. This combination of DTW and GMM provides a robust framework for unsupervised mode recognition in coupled flickering flames. In addition, this study compared two other unsupervised classification methods: the *k*-Shape classifier and the GMM classifier. The two conventional classifiers have similar procedures with the above DTW-GMM. Therefore, their structures are briefly illustrated in Fig. 5 and Fig. 6, respectively. Despite its capabilities for temporal information analysis, the DTW algorithm incurs high computational costs when applied to large datasets. Consequently, various efficient DTW variants, such as FastDTW [42], have been developed. In this study, to address the computational demands of DTW in large datasets, the nonlinear reduced-order model, VAE, is employed to map the high-dimensional flame data onto a two-dimensional latent space, significantly reducing computational overhead without excessive resource consumption.

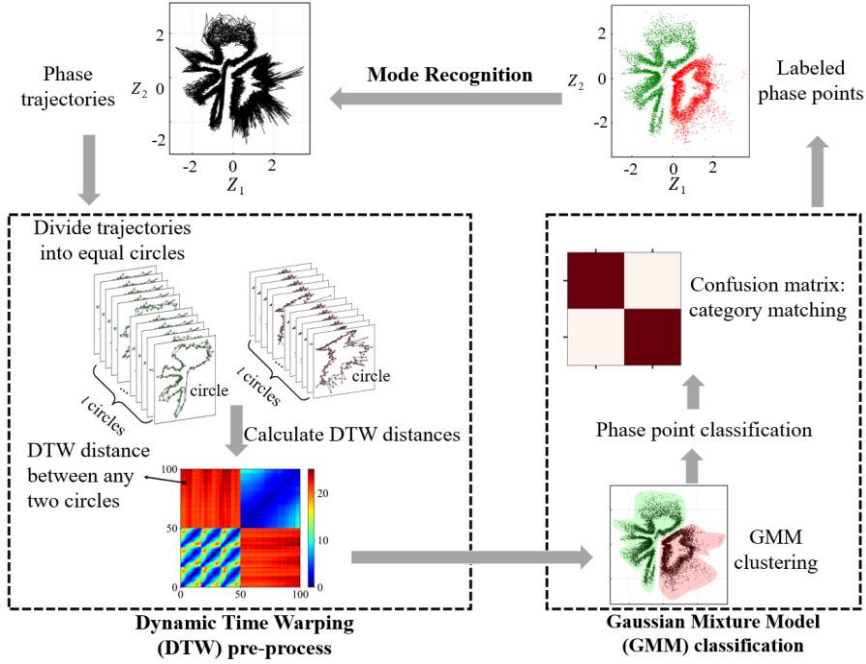

Fig. 4. Data flow through the proposed unsupervised GMM-DTW classifier modules: DTW alignment distance calculation and GMM clustering.

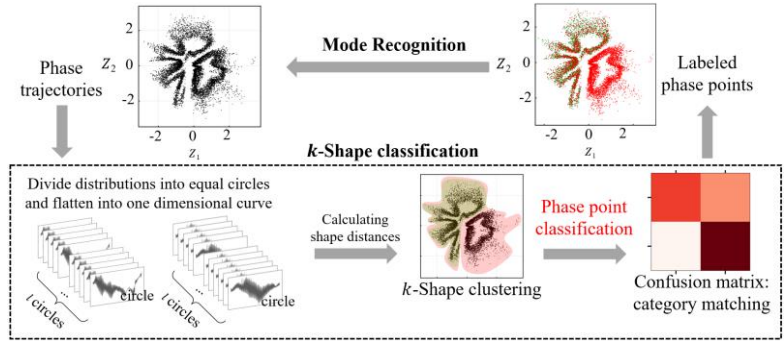

Fig. 5. Data flow analysis of comparison unsupervised $k$-Shape classifier module for time series: shape distances and clustering assignment.

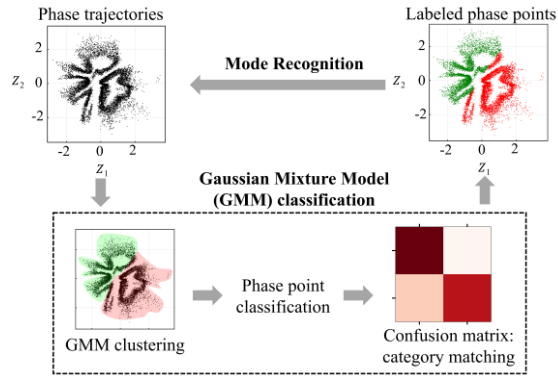

Fig. 6. Data flow analysis of comparison unsupervised GMM classifier module for time series.

### 2.4 Performance Metrics of Classifier Models

To assess the model recognition performance of the above models, the confusion matrix is a comparative analysis of the model's predictions regarding positive or negative outcomes against the actual class labels of the respective samples. The four key parameters are true positive (TP), false positive (FP), true negative (TN), and false negative (FN) values. Besides, a few indicators, including accuracy (AC), precision (PR), recall (RC, also known as True Positive Rate (TPR)), F1 score, false alarm rate (FAR, also known as False Positive Rate (FPR)), area under the curve (AUC), and geometric mean score (GMS), are widely used to evaluate the efficacy of classifier models [43]. The formulas of these performance indicators are shown as follows:

$$\text{Accuracy: } AC = \frac{TN+TP}{TN+TP+FN+FP} \tag{18}$$

$$\text{Precision: } PR = \frac{TP}{TP+FP} \tag{19}$$

$$\text{Recall: } RC\ (TPR) = \frac{TP}{TP+FN} \tag{20}$$

$$\text{F1 Score: } F1 = \frac{2 \times PR \times RC}{PR+RC} \tag{21}$$

$$\text{False Alarm Rate: } FAR\ (FPR) = \frac{FP}{FP+TN} \tag{22}$$

$$\text{Area Under the Curve: } AUC = \int_0^1 TPR(FPR)dFPR \tag{23}$$

$$\text{Geometric mean score: } GMS = \sqrt[N]{\prod_{i=1}^{N}(PR_i \times RC_i \times F1\ Score_i)} \tag{24}$$

where $N$ represents the number of classes.

Larger values of AC, PR, RC, F1 Score, AUC, and GMS indicate better model performance, but a small FAR value is desirable. These indicators individually evaluate various facets of the classification model, all of them offering a more comprehensive assessment. Specifically, the AC value scrutinizes the overall correctness of predictions; the PR value emphasizes the accuracy of positive class predictions; the RC value concentrates on the ability to capture positive instances; the F1 score considers a balance between precision and recall; the FAR value evaluates the degree of misclassification for the negative class. For binary classification problems, the AUC value measures the ability of the model to distinguish between classes, with a higher value indicating better discrimination, particularly for binary classification tasks. The GMS value calculates the geometric mean of performance metrics (AC, PR, RC, and F1 Score) for each class, providing a balanced assessment of the model's performance across all categories. This is especially crucial in multi-class classification problems.

Since the dynamical mode recognition of coupled flickering flame systems is a multi-class classification problem, the macro-averaged indicator of multi-class classification can be obtained as the mean value of these performance indicators:

$$Macro(AC, PR, RC, F1\ Score, FAR, GMS) = \frac{1}{N}\sum_{i=1}^{N}(AC, PR, RC, F1\ Score, FAR, GMS)_i \tag{25}$$

where $N$ represents the number of classes, $(AC, PR, RC, F1\ Score, FAR, GMS)_i$ is the $i$ th $(AC, PR, RC, F1\ Score, FAR, GMS)$ classification value.

### 3. Numerical Simulations of Flickering Flames and Data Collection

Flame flickering is a prevalent physical phenomenon observed in diffusion flames, premixed flames, and partially premixed flames. Many previous studies [44-47] have found the flicker of diffusion flames is a self-exciting flow oscillation, rather than the results of an externally forced vibration or the alternate flame extinction and re-ignition. The deformation, stretching, or even pinch-off of the flame surface results from the formation and evolution of the toroidal vortices. We utilize this kind of unsteady periodic flame as a prototype of the complex interaction of flame and vortex in many industrial devices, such as gas-stove burners [48], turbomachine rotors [49], can-annular combustors in gas-turbine engines [50, 51]. As shown in Fig. 7(a), the present flame configurations are single, dual, and triple flickering flames, where the Bunsen burner is a cube column of $D \times D \times 3D$ with a fuel inlet of methane at a velocity of $U_0$ and $l$ is a gap distance between each two identical flames. The detailed descriptions of numerical schemes, parameters, and validations of present flame simulations can be referred to the Supporting Material and the previous studies [52-55].

In a single flickering flame, the deformation, stretching, or even pinch-off of the flame surface results from the formation and evolution of the toroidal vortices. Taking a single flickering flame as an example, the flame flicker is illustrated briefly here. Fig. 7(b) shows the flame shape and the vorticity around the flame during one cycle of $\Delta t \approx 0.1\text{s}$. In physics, the vorticity accumulation inside the toroidal vortex causes the flame deformation; then the rapture of the flame neck can pinch off the flame; finally, the upper portion of the separated flame is convected downstream with the toroidal vortex. Likely, a new circle will start again. Details on the connection between flame flicker and vortex evolution can be referred to [56].

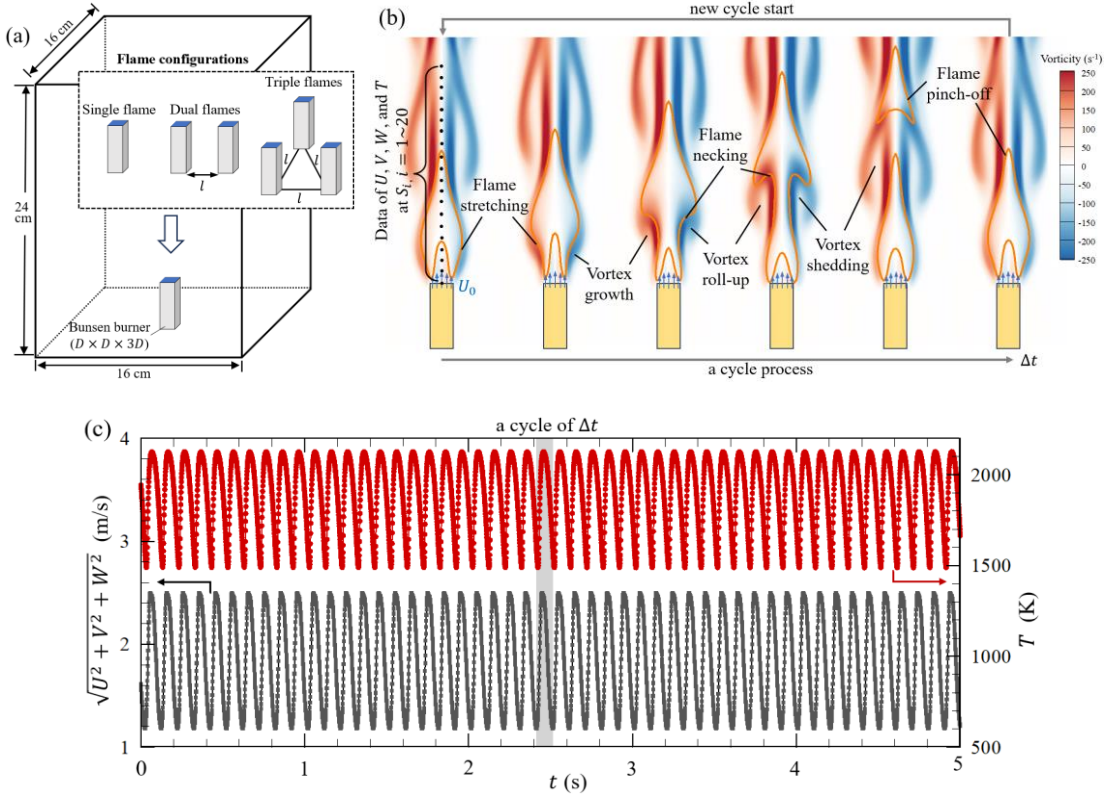

Fig. 7. (a) Computational setup of single-, dual-, and triple-flame systems. The Bunsen burner is a cube column of $D \times D \times 3D$, where the top side is set as a fuel inlet of methane with a velocity of $U_0$. The dual-flame system contains two identical Bunsen burners with a gap distance of $l$, while three burners are arranged in an equilateral triangle with the same distance of $l$ for the triple flames system. (b)

Dynamical behavior of single flickering buoyant diffusion flame: physical features (flame shape and vorticity) during a cycle process of $\Delta t$. (c) Time-varying velocity magnitude of $\sqrt{U^2 + V^2 + W^2}$ (m/s) and temperature of $T$ (K) within 5 s. The data are obtained from the sixth one of 20 sensors arranged equally along the flame center, as shown in Fig. 7(b).

To train and test our proposed models, datasets are obtained from 80 features including 4 variables of velocity components ($U$, $V$, and $W$) and temperature $T$ of 20 sensors, which were collected at a sample frequency of 1000 Hz along each flame centerline for the dimensionality reduction analysis. To further display the periodic oscillation, velocity magnitude and temperature of the sixth sensor during 5 s (about 50 cycles) are plotted in Fig. 7(c). The sampling frequency is much higher than the flickering frequency so that small time-scale information of coupled flames can be extracted for our proposed models. In the present study, all simulation cases of dual- and triple-flame systems with flame parameters and dynamical modes are listed in Table S2.

When the gap distance $l$ between each two flames is varied, the dual-flame and triple-flame systems exhibit various dynamic modes, as mentioned in Section 1. Our previous works [52, 53] numerically investigated and theoretically analyzed the dynamical behaviors of dual and triple buoyant diffusion flames. It is key to understand the flickering mechanism of a single buoyant diffusion flame, as shown in Fig. 7. A brief introduction to coupled flickering flames is given here. From the perspective of vortex dynamics, we attempted to understand the anti-phase and in-phase flickering in a dual-flame system [52], where the interaction between vortex rings of the two flames plays an important role. As shown in Fig. 8(a-b), the in-phase (IP) mode appears as the dual flames flicker synchronously with a negligible phase difference, while the anti-phase (AP) mode appears as the flames alternatively flicker with a fixed phase difference of half-period. Physically, the transition of the vortical structures from symmetric (in-phase) to staggered (anti-phase) can be interpreted as being similar to the mechanism causing flow transition in the wake of a bluff body and forming the Karman vortex street.

Similarly, the four distinct dynamical modes (in-phase, death, rotation, and partially in-phase) of triple flickering buoyant diffusion flames in an equilateral triangle arrangement can be interpreted based on vortex interaction [53]. Specifically, the in-phase (IP) mode with a phase difference of $0\pi$ in Fig. 8(c) is caused by the periodic shedding of the trefoil vortex formed by the reconnection of three toroidal vortices; the death mode with the flickering suppression (very small oscillation amplitude) in Fig. 8(d) is due to the suppression of vortex shedding at small Reynolds numbers; the rotation mode with a fixed phase difference of $2\pi/3$ in Fig. 8(e) appears as three toroidal vortices alternatively shed off with a constant phase difference; the partially in-phase (PIP) model (in-phase of two flames but anti-phase to another one) with $0\pi$ and $\pi$ in Fig. 8(f) is caused by the vorticity reconnection of two toroidal vortices leaving another one shedding off in the anti-phase way. Therefore, studying coupled flickering flames can be very well to establish a bridge between the vortex dynamics and the nonlinear dynamics of the couple flame system. However, conventional knowledge-based approaches have limits to identifying larger dynamical systems of multiple flames.

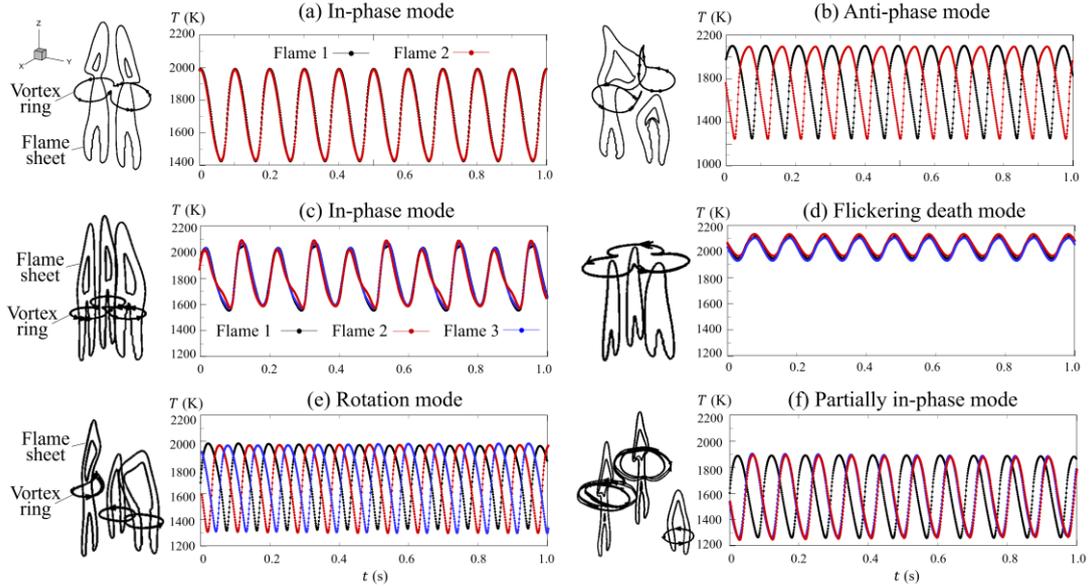

Fig. 8. Various dynamical modes induced by vortex interaction and the temperature variation with time of dual-flame and triple-flame systems: two identical flickering flames [52] in (a) in-phase (IP) mode and (b) anti-phase (AP) mode; three identical flickering flames [53] in (c) in-phase (IP) mode, (d) flickering death mode, (e) rotation mode, and (f) partially in-phase (PIP) mode.

## 4. Results and Discussion
### 4.1 Dimensionality Reduction of High-dimensional Spatiotemporal Flame Dataset

As a benchmark comparison of the nonlinear VAE and the linear PCA models, Fig. 9(a) and 9(b) both reveal repeated limit cycles in the continuous phase trajectory in the two-dimensional latent space. However, with the same dimensionality compression ratio, the mean squared error (MSE) of VAE reconstruction of about 0.2 is smaller than 0.4 of the first two-eigenvalue reconstruction of PCA. In other words, the linear PCA needs at least the first five eigenvalues to be comparable to the nonlinear VAE construction. More details on the MSE error of single-, dual-, and triple-flame systems are given in Table S3 of the Supporting Material. It should be noted that the complexities of the compared methods are not the same, and the goal of this comparison is to assess how different model architectures affect the quality of the reconstructions. The present results show that the reconstruction performance of VAE becomes increasingly superior to PCA with the increasing complexity of combustion systems.

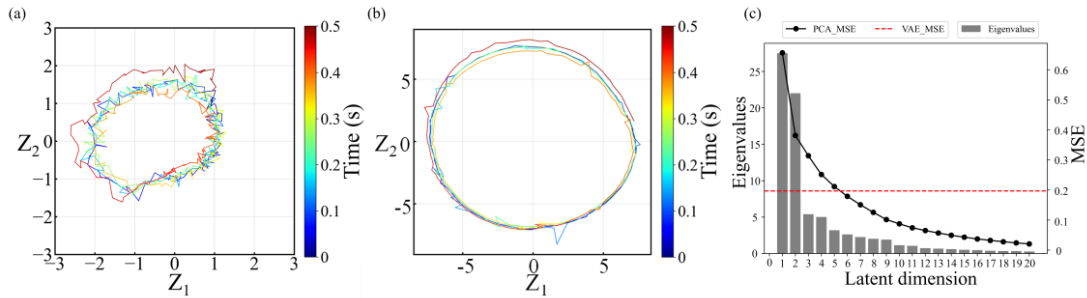

Fig. 9. Phase trajectory of the nonlinear oscillation of single buoyant diffusion flame via (a) the latent space of VAE and (b) the first two-eigenvalue space of PCA. (c) MSE errors of the present VAE with two latent dimensions and the PCA across various latent dimensions.

Similarly, Fig. 10 compares the dimensionality reduction analysis of VAE and PCA models for dual-flame and triple-flame systems. The phase distributions in the two-dimensional latent space of VAE and PCA are depicted in Fig. 10(a-b) and Fig. 10(d-e), where the different dynamic modes are represented by different colors. The present results show that the shapes of phase trajectories from different scenarios are notably distinct. For the dual-flame system featuring two dynamical modes, both VAE and PCA exhibit distinct separation of phase points representing different modes in the latent space, as shown in Fig. 10(a) and Fig. 10 (b). Fig. 10 (c) further illustrates that the nonlinear VAE achieves significantly lower reconstruction errors compared to PCA at the same dimensionality compression ratio, with PCA requires at least the first 17 eigenvalues to achieve a comparable reconstruction error in the dual-flame system.

However, with the increasing complexity of combustion systems, such as the triple-flame system exhibiting multiple modes, traditional linear methods like PCA often fail to distinctly clarify the various dynamic modes due to their linear constraints. In such cases, PCA may reveal an overlap of phase points representing different modes, blurring the distinction between them, as shown in Fig. 10(e). Conversely, VAE demonstrates the ability to maintain a clear separation between modes, as shown in Fig. 10(d), owing to its capacity to capture nonlinear relationships. Fig. 10(f) demonstrates that in the triple-flame system, the nonlinear VAE consistently achieves lower reconstruction errors compared to PCA at the same dimensionality compression ratio. To achieve a comparable reconstruction error, PCA requires retaining at least the first 18 eigenvalues, highlighting VAE's superior performance in handling the more complex dynamics of the triple-flame system. The comparison shows that this classifying capability of VAE could be a promising approach for tackling complex combustion systems and elucidating the distinct dynamics inherent within them, particularly for the tasks related to recognizing modes in complex flame systems.

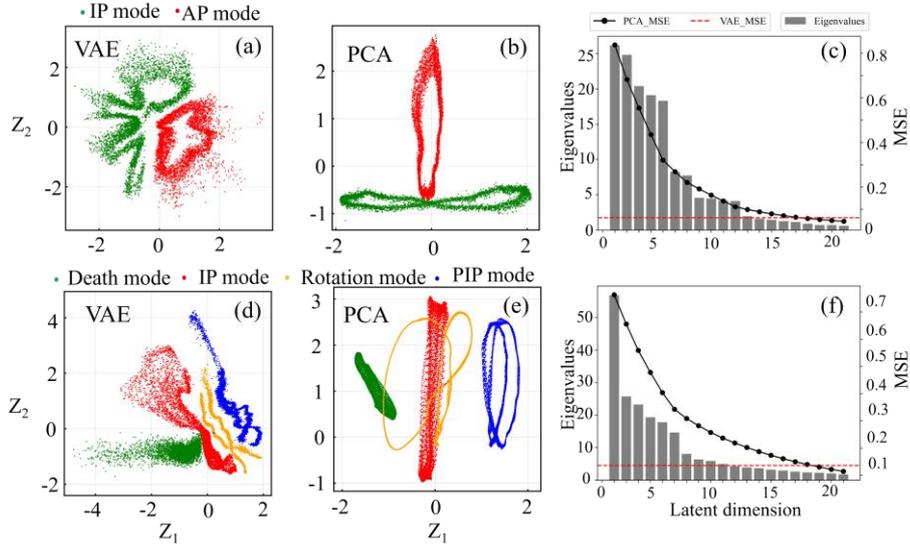

Fig. 10. Phase distributions of the coupled flickering flames via VAE and PCA: (a-c) dual-flame system and (d-f) triple-flame system. Different colors of phase points represent the different dynamic modes, including in-phase (IP) and anti-phase (AP) modes in a dual-flame system and in-phase (IP), death, rotation, and partially in-phase (PIP) modes in a triple-flame system.

Hereto, we have compared the dimensionality reduction of the nonlinear VAE and the linear PCA models for high-dimensional spatiotemporal dual-flame and triple-flame systems. We found that the VAE performance is significantly superior to that of PCA in classifying dynamical modes of nonlinear combustion systems. Firstly, the linearity makes PCA-based ROM suffer from performance degradation in nonlinear problems, requiring an excessive number of modes compared with nonlinear ROM model VAE for the same reconstruction accuracy. Moreover, when projecting data into a lower-dimensional latent space, PCA may blur the distinction between different modes by overlapping phase points, resulting in poorer performance in subsequent recognition tasks. While VAE stands out for its capability to effectively capture nonlinear relationships, thus maintaining a clear separation between modes.

**4.2 Wasserstein-distance-based Supervised Classification**

In supervised classification, the training data includes 50 cycles with well-labeled data for each mode. To ensure model robustness and effectiveness, 20% of this training dataset is set aside for validation purposes. Additionally, there is a prediction dataset for each mode, consisting of 20 cycles of data. It's noteworthy that all cycles within both the training and prediction datasets share the same time duration. For each cycle, WD values are computed in comparison to each benchmark case, which consists of well-labeled data. Specifically, Fig. 11(a) shows that the prediction dataset for a double flames system with two dynamical modes comprises a total of 40 cycles of data. These cycles are individually compared to the in-phase and anti-phase benchmark datasets to calculate WD values. A smaller WD value indicates greater similarity between the two distributions. For instance, the WD values for the first 20 cycles with the anti-phase dataset are relatively small, leading to the identification of these cycles as data generated by the anti-phase mode. Consequently, the subsequent 20 cycles are recognized as data produced by the in-phase mode. This accurate identification process is similarly applicable to the prediction dataset of a triple-flame system, as shown in Fig. 11(b), facilitating the recognition of data corresponding to the four modes.

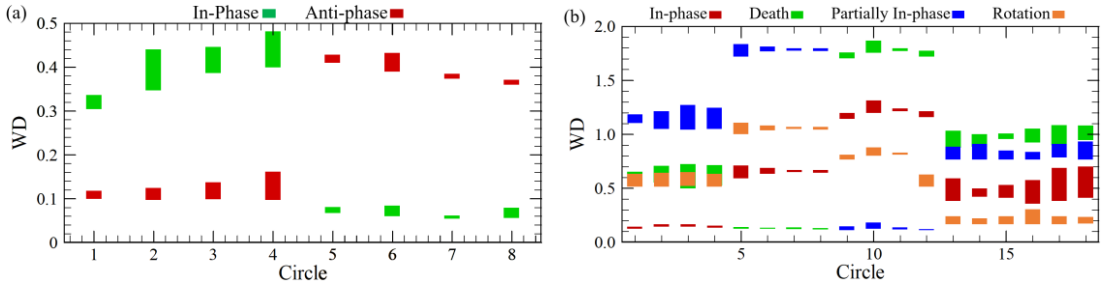

Fig. 11. Wasserstein distance (WD) values of the (a) dual- and (b) triple-flame systems. Each circle represents 5 cycles of data collected from the combustion systems. Different colors denote the WD values between the prediction data and the benchmark of the different dynamic modes including in-phase and anti-phase modes in a dual-flame system and in-phase, death, rotation, and partially in-phase modes in a triple-flame system.

**4.3 GMM-DTW-based Unsupervised Classification**

However, industrial processes acquire a thoroughly labeled dataset, which is frequently both costly and challenging. Therefore, this study integrates the intrinsic spatial-temporal characteristics of flame data and utilizes the unsupervised machine learning clustering approach, GMM-DTW, for flame dynamical mode recognition. Also, the training dataset comprises 50 cycles of unlabeled data for each mode, 20% of this training dataset is designated for validation purposes to uphold the robustness and effectiveness, and each mode has a prediction dataset containing 20 cycles.

Figure 12 shows the clustering results of GMM-DTW, GMM, and *k*-Shape clustering methods in the dual-flame and triple-flame system, where phase points representing different categories are visually distinguished by using different colors. Based on the time trajectories of the phase points, our proposed GMM-DTW classifier employs the DTW algorithm to capture periodicity. The GMM clustering method directly utilizes phase point distributions. The *k*-Shape clustering method is based on the shape distances of flattened one-dimensional sequences. Comparing the results of three clustering methods with true labels in the dual-flame system in Fig. 12(a-d) and the triple-flame system in Fig. 12(e-h), we found that the proposed GMM-DTW algorithm has a better classification than others. The GMM and *k*-Shape methods both exhibit some deviations in classifying accurate modes. Specifically, the GMM-DTW in the dual-flame system accurately identifies each phase point for the corresponding mode (clear separation of two modes in the latent space), as shown in Fig. 12(b). However, GMM and *k*-Shape both have poor performance. For the triple-flame system with more complex dynamics, GMM-DTW still discerns the corresponding mode of each phase point exactly, while GMM and *k*-Shape results deviate significantly from the true distribution of the four distinct modes.

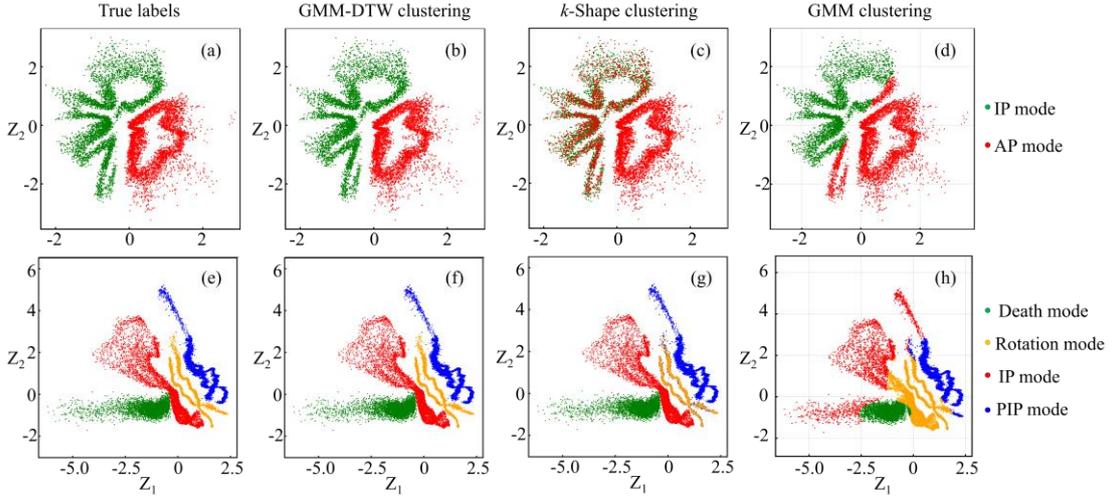

Fig. 12. Clustering results of the true labels, GMM-DTW, *k*-Shape, and GMM clustering for (a-d) dual-flame and (e-h) triple-flame systems. Different colors of phase points represent the different dynamic modes, including in-phase (IP) and anti-phase (AP) modes in the dual-flame system and in-phase (IP), death, rotation, and partially in-phase (PIP) modes in the triple-flame system.

To effectively evaluate the performance of the classifier models, this study utilizes a confusion matrix and a set of performance evaluation indicators, as introduced in detail in Section 2.4. We aim to establish an unsupervised model for effectively recognizing various dynamical modes of multi-flame systems. Specifically, the detailed indicator results of the VAE-GMM-DTW, VAE-GMM, and VAE-*k*-Shape classification models for the double and triple flame systems are summarized in Table S4 of the Supporting Material. Their recognition performance is analyzed as follows.

Figure 13 shows the confusion matrix of the present mode recognition, where the predictive performances of the three models across different classes are provided in detail. For dual-flame system, the proposed VAE-GMM-DTW model predicts very well, while VAE-GMM and VAE-*k*-Shape models more or less have inaccurate predictions. The same conclusions can be drawn in the triple-flame system with more complex dynamics (four dynamical modes) in flame and vortex interaction. Therefore, the above performance indicators can give us a side confirmation of the performance of the three classification models across various dynamical modes of complex flame systems.

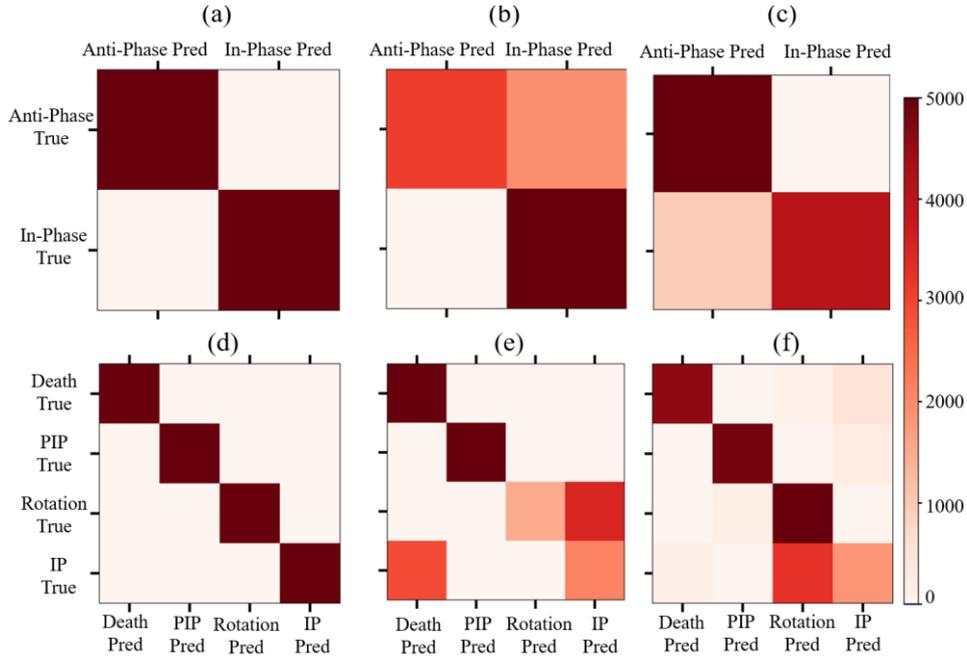

Fig. 13. Confusion matrix of the VAE-GMM-DTW, VAE-*k*-Shape, VAE-GMM clustering approaches for (a-c) in-phase and anti-phase modes of the dual-flame system and (d-f) in-phase (IP), death, rotation, and partially in-phase (PIP) modes of the triple-flame system.

For binary classification problems in the dual-flame system, we use the AUC metric to evaluate the overall performance of the three classification models, as shown in Fig. 14. This metric measures the model performance for accurately distinguishing between positive and negative samples, thereby effectively addressing the challenge of imbalanced data. To further quantify the classification performance and analyze the impact of elaborating the design of each module in the proposed model architecture, the AC, PR, RC, F1 score, FAR values, and GMS in the double flame system and the macro-AC, macro-PR, macro-RC, macro-F1 score, macro-FAR values, and macro-GMS in the triple-flame system are depicted in Fig. 15. For these flame systems, the present results show that the overall mean accuracy, precision, recall, F1 Score, and GMS values of the proposed VAE-GMM-DTW model are 1, while the FAR value is 0. It indicates that the proposed model demonstrates excellent performance in the classifying tasks of these flame systems.

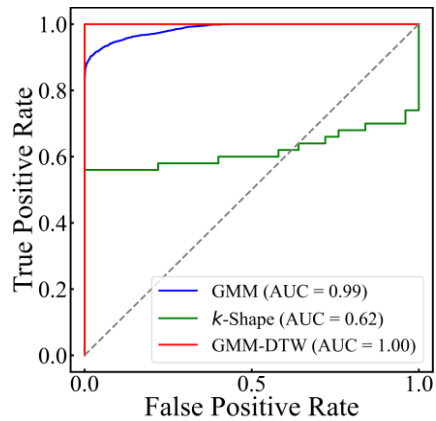

Fig. 14. Overall categorized performance of the VAE-GMM-DTW, VAE-GMM, and VAE-*k*-Shape approaches in the dual-flame system using the AUC.

To verify the effect of both the temporal and spatial characteristics of phase points on dynamic mode recognition in complex flame systems, we conduct a comparative analysis from two perspectives. First, compared to the VAE-GMM model with just incorporating spatial information, the proposed VAE-GMM-DTW consistently outperforms it in both binary classification (dual-flame cases) and multi-class classification (triple-flame cases). Positive indicators, such as AC, PR, RC, F1 score, and GMS values of the proposed model are higher and the negative indicator FAR is significantly lower, which indicates a superior model performance. Therefore, the present results further highlight the advantage of VAE-GMM-DTW and emphasize the temporal characteristic of phase points plays an important role in correctly recognizing dynamical modes in complex flame systems. Second, the proposed VAE-GMM-DTW also demonstrates better performance than the VAE-$k$-Shape (temporal information is only focused) in both binary and multi-class classifications. Similarly, considering the two-dimensional spatial characteristics of phase points in the VAE-GMM-DTW enhances the accurate recognition of dynamical modes in complex flame systems.

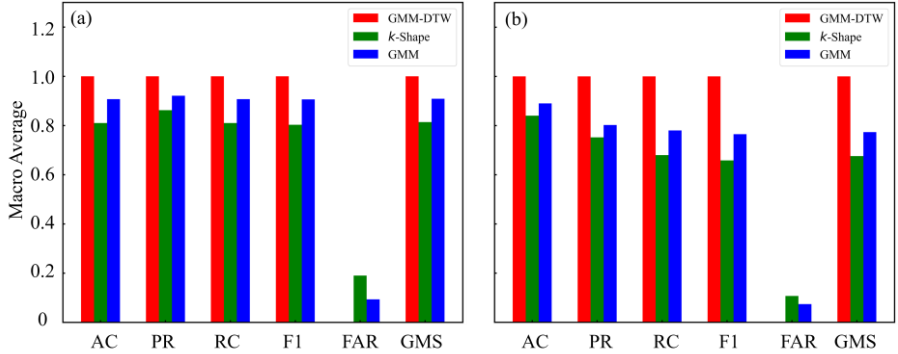

Fig. 15. Overall categorized performance of the VAE-GMM-DTW, VAE-GMM, and VAE-$k$-Shape approaches in (a) dual-flame and (b) triple-flame systems.

To further validate the performance of the proposed VAE-GMM-DTW in recognizing each subclass in a multi-classification task, we obtain the AC, PR, RC, F1 score, FAR, and GMS values for each category in the triple-flame system with a multi-classification task. The corresponding values of these indicators are provided in Table S5 of the Supporting Material. According to the results in Fig. 16, the subclass categorized performance of VAE-GMM-DTW is consistent with its overall categorized performance, as the AC, PR, RC, F1 score, and GMS indicators are all 1 for the four modes. These values are significantly higher than those achieved by the GMM and $k$-Shape models. In addition, the negative indicator FAR of VAE-GMM-DTW is 0, notably lower than that of the GMM and $k$-Shape. Therefore, we can conclude that the proposed VAE-GMM-DTW model not only delivers optimal overall classification performance in multi-classification scenarios but also excels in the classification of each subclass.

In summary, the proposed VAE-GMM-DTW model effectively captures temporal information from dynamical flame systems using the GMM-DTW clustering approach. The results underscore the robust performance of the proposed model in both binary classification (dual-flame system) and multi-class classification (triple-flame system), emphasizing its ability to achieve superior outcomes of dynamical mode recognition, characterized by both high accuracy and minimal false alarm rates. The present study indicates the potential of the proposed approach to address the challenges in complex coupled flame oscillator systems.

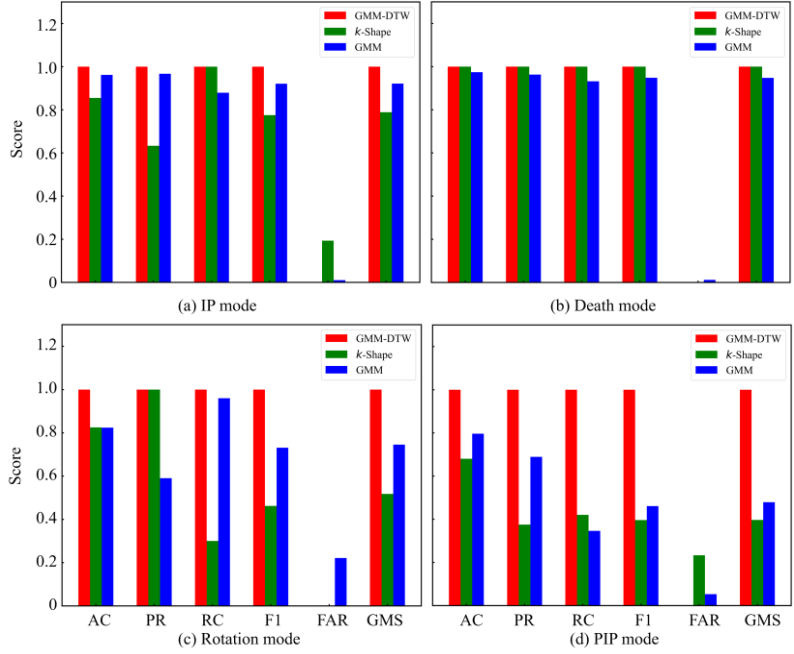

Fig. 16. Categorized performance of the proposed VAE-GMM-DTW approach for (a) in-phase (IP), (b) flickering death, (c) rotation, and (d) partially in-phase (PIP) modes in the triple-flame system.

**4.4 Reliability Analysis of the Proposed Supervised and Unsupervised Classification Approaches**

To thoroughly evaluate the stability of the proposed approaches, we conduct a sensitivity analysis focusing on key parameters such as batch size and learning rate. The performance of both the WD-based supervised classification and GMM-DTW-based unsupervised classification methods are assessed by systematically varying these key parameters. The sensitivity analysis includes the testing batch sizes of 16, 32, 64, and 128, as well as the learning rate of $1 \times 10^{-4}$, $5 \times 10^{-4}$, $1 \times 10^{-3}$, and $5 \times 10^{-3}$.

As depicted in Fig. 17 and Fig. 18, the performance metrics, including WD, AC, PR, RC, F1 Score, FAR, and GMS values of WD-based supervised classification and GMM-DTW-based unsupervised classification, remain consistent across all tested configurations. These results also clearly show that our proposed GMM-DTW-based unsupervised classification demonstrates superior stability and performance, compared to the VAE-*k*-Shape and VAE-GMM. Specifically, the GMM-DTW maintains high AC, PR, RC, F1 Score, FAR, and GMS across all tested batch sizes and learning rates, while VAE-*k*-Shape and VAE-GMM exhibit notable fluctuations in these metrics. For instance, the accuracy of the VAE-*k*-Shape is significantly improved when the batch size increases from 16 to 128. The VAE-GMM has inconsistent precision and recall values at different learning rates. This fluctuation of indicator values indicates that compared to the GMM-DTW, the VAE-*k*-Shape and VAE-GMM are more sensitive to parameter changes, leading to less consistent performance. This enhanced performance of the GMM-DTW can be attributed to the utilization of the DTW algorithm to capture temporal information in the latent space, which is crucial for the accurate recognition of dynamical modes in the present flame systems. Again, these findings confirm the robustness of our proposed approach, which is not only effective but also resilient to parameter variations, ensuring reliable recognition of dynamical modes under various conditions.

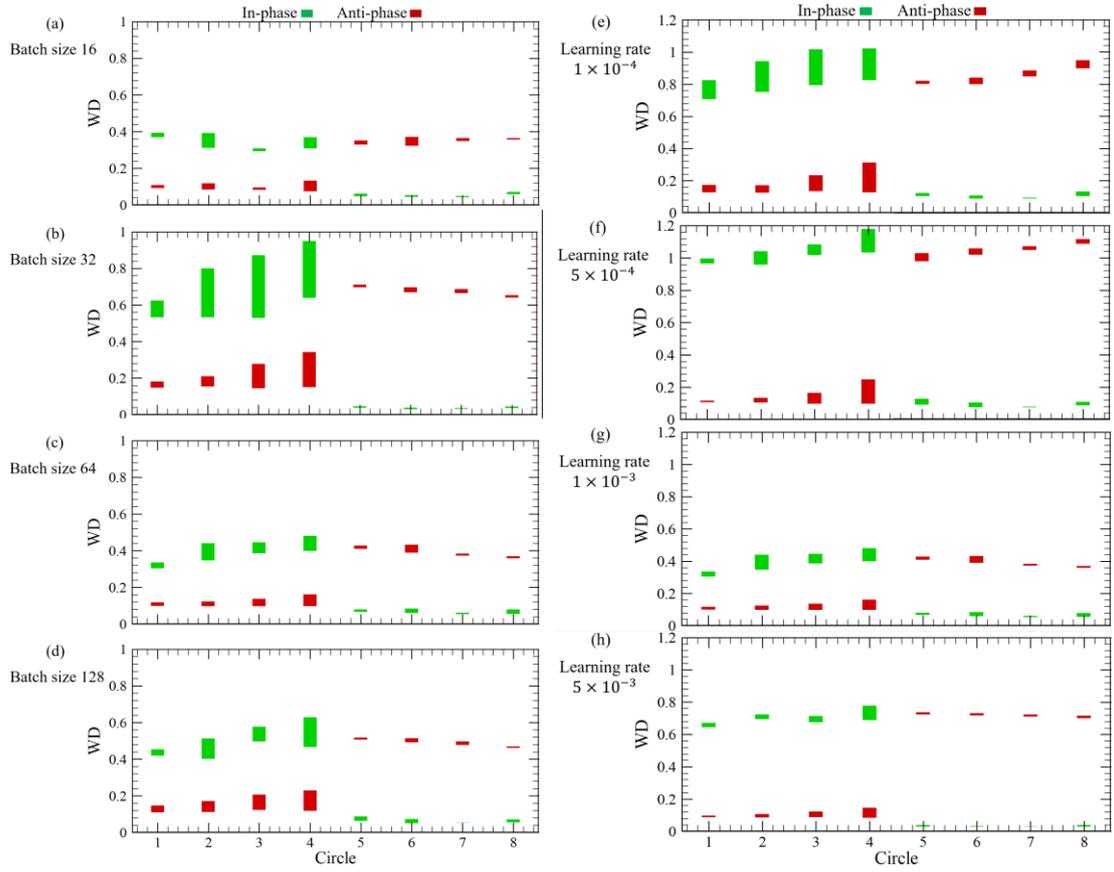

Fig. 17. Sensitivity analysis of the batch size (a-d) and learning rate (e-h) for WD-based supervised classification approach for in-phase and anti-phase modes in the dual-flame system.

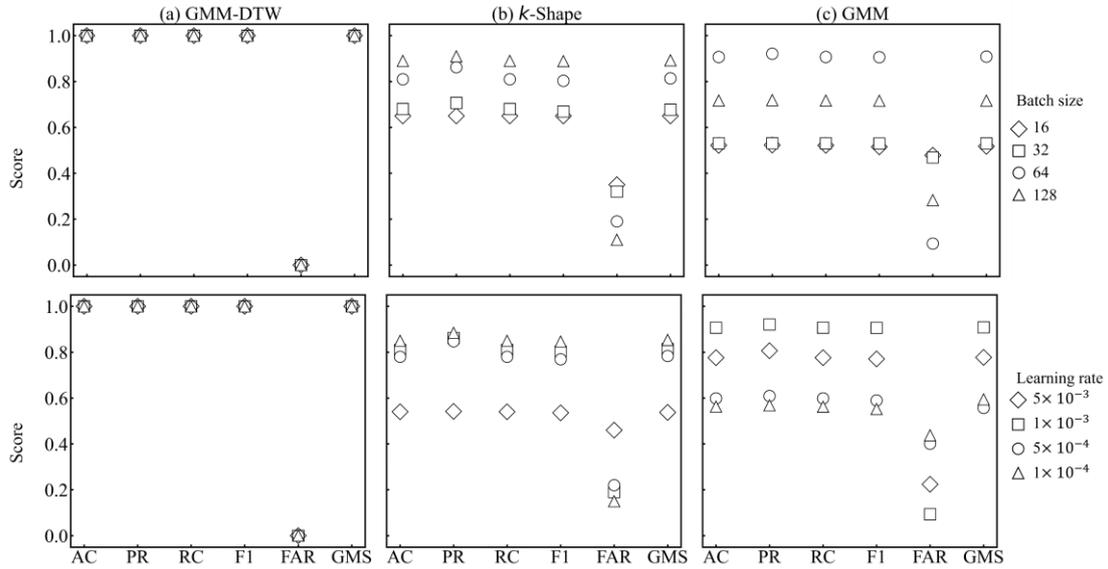

Fig. 18. Sensitivity analysis of batch size and learning rate for the VAE-GMM-DTW, VAE-GMM, and VAE-$k$-Shape approaches for in-phase and anti-phase modes in the dual-flame system.

To further evaluate the robustness of our proposed approaches, we conduct additional case studies involving dual-flame and triple-flame systems under different operating conditions (i.e., Dual B cases and Triple B cases in Table S2), providing a comprehensive performance evaluation of the present supervised and unsupervised classifiers. As illustrated in Fig. 19 and Fig. 20, both the WD-based supervised classification and GMM-DTW-based unsupervised classification consistently demonstrate significant recognition performance across different conditions in both dual-flame and triple-flame systems. These methods reliably identified dynamical modes, further validating their robustness under varying operating conditions. Conversely, the VAE-$k$-Shape and VAE-GMM showed more variable performance under similar conditions. Specifically, when subjected to different conditions within the triple-flame system, the VAE-$k$-Shape exhibited notable variability in the six-evaluation metrics with precision dropping significantly in response to new operational scenarios. Similarly, the VAE-GMM displays inconsistent precision and recall values across different conditions in the triple-flame system, highlighting its sensitivity to condition changes. The present results indicate that the VAE-$k$-Shape and VAE-GMM are less stable and robust, compared to the GMM-DTW-based method, particularly under varying operating conditions.

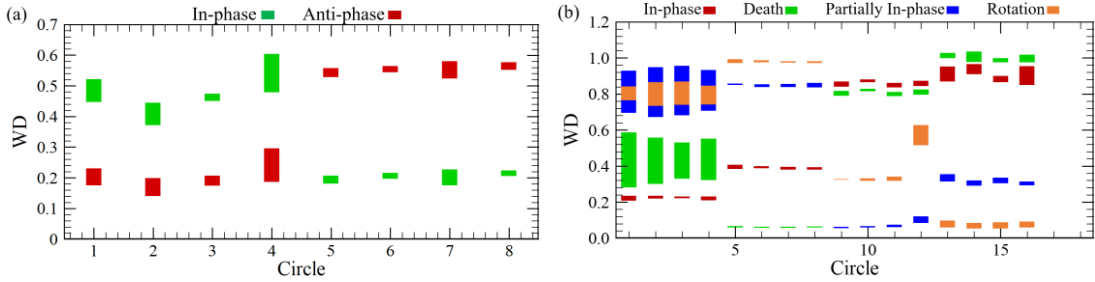

Fig. 19. Robustness analysis of the proposed VAE-WD for (a) Dual B cases in the dual-flame system and (b) Triple B cases in the triple-flame system.

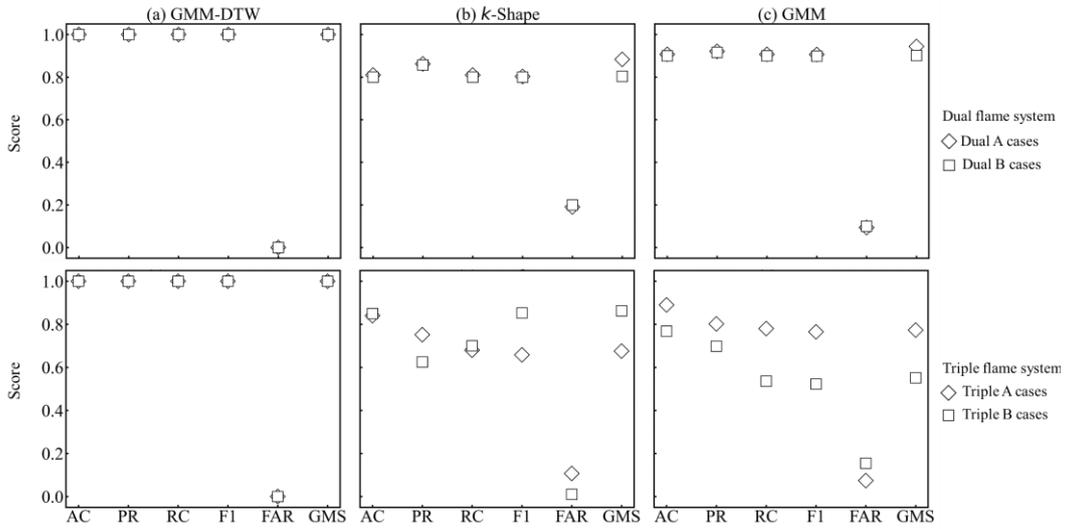

Fig. 20. Robustness analysis of the (a)VAE-GMM-DTW, (b)VAE-$k$-Shape, and (c) VAE-GMM unsupervised approaches for Dual A and B cases in a dual-flame system and Triple A and B cases in a triple-flame system under varying conditions.

As a deep learning approach, the VAE model inherently offers robust scalability owing to its flexible framework and powerful potential for learning latent representations. The VAE model adeptly captures intricate data distributions by projecting high-dimensional data into the low-dimensional latent space, dealing with varying dataset sizes and complexities in model architecture. This flexibility not only accommodates diverse data characteristics but also supports scalable model architectures, making VAE suitable for applications beyond the focus of the present study. For instance, recent applications of VAE-based methodologies in mode recognition within circular arrays of flame oscillators [22] and anomaly detection in industrial steam turbine power systems [57] accentuate the adaptability and scalability in tackling complex real-world challenges. This expanded view highlights the VAE model as a versatile tool capable of scaling effectively across different domains and data environments, ensuring robust performance in diverse applications.

The present investigation also delves into the optimal conditions for deploying these approaches. Specifically, the proposed approaches perform most effectively in environments, where the system dynamics are relatively stable and well-defined, such as in laminar flow conditions. In these scenarios, the flame structures exhibit consistent vortex patterns, allowing our VAE-based model to capture and represent the underlying features accurately. The absence of turbulence ensures that the extracted features remain coherent and interpretable, leading to more reliable mode recognition. In turbulent environments, where chaotic and unpredictable behaviors are prevalent, the complexity of the flame dynamics increases significantly. This poses challenges for feature extraction and may require further refinement of the proposed models or additional pre-processing steps to manage the noise and variability.

It is worth noting that VAE-based approaches for dimensionality reduction and classification require careful management to avoid potential pitfalls. Improper model design or training can lead to posterior collapse, where the KL divergence dominates over the reconstruction term, limiting the latent space's ability to capture meaningful variations of data. Effective hyperparameter tuning, including adjustments to learning rate, batch size, and neural network architecture, is crucial for ensuring the VAE converges to optimal solutions without sacrificing the diversity of learned latent representations. Some model variants, like $\beta$-VAE, aim to mitigate posterior collapse but introduce additional complexity, necessitating thorough tuning and analysis for optimal performance in practical applications.

## 5. Conclusions

Nonlinear dynamics of coupled oscillators is a long-lasting problem in the science of complex systems. The work presented in this study takes a baby step toward the objective of studying the collective dynamical behaviors of coupled flame oscillators made of laminar flickering diffusion flames. Different from the conventional approaches of nonlinear dynamics and complex systems theories, we adopted machine learning-based approaches to provide new perspectives to recognize and classify various dynamical modes of the coupled flame oscillators.

Specifically, a VAE-WDC supervised classification model and a VAE-GMM-DTW unsupervised classification model for binary classification and multiclass classification identification of coupled flame oscillator systems are established in this study. The data are collected from the fully validated simulations of diverse dynamical modes in single-, dual-, and triple-flame systems. These simulations involve solving the buoyancy-driven flickering flames using an unsteady, three-dimensional, low-Mach, and variable-density chemically reacting flow with a simplified reaction mechanism. Each dataset comprises a time series with 80 features from each flame, where four variables ($U$, $V$, $W$, and $T$) were sampled at 1000 Hz for 5 seconds (approximately 50 flickering periods) along the central axis of each flame.

Based on supervised and unsupervised methods of deep learning, we developed a robust and comprehensive framework for dynamical mode recognition of coupled flame oscillators, which has the potential for larger combustion systems involving real turbulent flames. The established nonlinear dimensional reduction VAE model is effective in projecting the high-dimensional data from multi-flame systems onto a 2-dimensional phase space defined by two latent variables, providing a data-driven feature extraction approach compared to expert knowledge-based methods. It holds the potential for overcoming the limitations of feature extraction through expert knowledge and for applications in high-complexity nonlinear dynamic systems.

Additionally, using deep-learning-based dimensionality reduction, we provided new ways of classifying these reduced spatiotemporal data of the coupled flame oscillators. For datasets with sufficient labels, we take the distribution of flame data in the phase space as the benchmark, then implement supervised mode recognition using the Wasserstein distance (WD) as a quantitative measure of proximity between two probability distributions in latent space. To address unlabeled time series data, we combine the dynamic time warping (DTW) and the unsupervised Gaussian mixture model (GMM) to achieve effective recognition of the dynamical mode of complex flame systems based on phase point trajectories. In this study, both supervised and unsupervised mode recognition approaches yield favorable classification results.

The present study demonstrates the first attempt at providing a robust and comprehensive framework for mode recognition of coupled flickering flames. By using the VAE model for non-linear dimensionality reduction, our approach can overcome the robustness and scalability limitations of expert-driven feature extraction in physical models [29], mapping high-dimensional flame data into a lower-dimensional space for dynamical mode recognition. This framework holds potential for applications with more complex combustion.

**Declaration of competing interest**

The authors declare that they have no conflict of interest.

**Acknowledgments**

This work is supported by the National Natural Science Foundation of China (No. 52176134) and partially by the APRC-CityU New Research Initiatives/Infrastructure Support from Central of City University of Hong Kong (No. 9610601).

**Supplementary material**

Supplementary material includes the details of the present VAE formulation, the numerical simulations, the reconstruction loss of training and validating VAE, and the evaluation results for single, dual-, and triple-flame systems.

**Author Contributions**

**Weiming Xu**: Conceptualization (equal); Data curation (equal); Formal analysis (equal); Investigation (equal); Methodology (lead); Visualization (equal); Writing – original draft (equal). **Tao Yang**: Conceptualization (equal); Data curation (equal); Formal analysis (equal); Investigation (equal); Simulation (lead); Visualization (equal); Writing – original draft (equal). **Peng Zhang**: Conceptualization (equal); Investigation (equal); Supervision (lead); Writing – review and editing (lead).